\documentclass{article}

\PassOptionsToPackage{numbers,compress}{natbib}

\usepackage[preprint]{neurips_2025}

\usepackage[utf8]{inputenc}
\usepackage[T1]{fontenc}

\usepackage{graphicx}
\usepackage{booktabs}

\usepackage{hyperref}       
\hypersetup{
    colorlinks=true,
    linkcolor={[rgb]{0.05,0.20,0.60}},
    citecolor={[rgb]{0.05,0.20,0.60}},
    urlcolor={[rgb]{0.05,0.20,0.60}}
}
\usepackage{url}
\usepackage{booktabs}
\usepackage{amsfonts}
\usepackage{nicefrac}
\usepackage{microtype}
\usepackage{xcolor}

\usepackage[accsupp]{axessibility}  
 \usepackage{multirow} 
\usepackage{wrapfig}

\usepackage{xspace}
\usepackage{subcaption}
\newcommand{\eg}{e.g.\xspace}
\newcommand{\ie}{i.e.\xspace}

\usepackage{orcidlink}

\newcommand{\vaename}{TrajLoom-VAE}
\newcommand{\modelname}{TrajLoom-Flow}
\newcommand{\benchmarkname}{TrajLoomBench}
\newcommand{\encodingname}{Grid-Anchor Offset Encoding}

\makeatletter
\newcommand{\reflike}[1]{\begingroup\color{\@linkcolor}#1\endgroup}
\makeatother

\begin{document}

\title{TrajLoom: Dense Future \\ Trajectory Generation from Video} 




\author{%
Zewei Zhang$^{1}$ \\
\texttt{zhanz561@mcmaster.ca}\And
Jia Jun Cheng Xian$^{2,3}$ \\
\texttt{anthony@ece.ubc.ca}\And
Kaiwen Liu$^{2,3}$ \\
\texttt{kaiwenliu@ece.ubc.ca}\AND
\hspace*{-0.8cm}Ming Liang$^{4}$ \\
\hspace*{-0.8cm}\texttt{liangming.elgoog@gmail.com}\And
\hspace*{0.8cm}Hang Chu$^{4}$ \\
\hspace*{0.8cm}\texttt{hang.chu@warpengine.ai}\AND
\hspace*{-0.8cm}Jun Chen$^{1}$\\
\hspace*{-0.8cm}\texttt{chenjun@mcmaster.ca}\And
\hspace*{0.8cm}Renjie Liao$^{2,3,5}$\\
\hspace*{0.8cm}\texttt{rjliao@ece.ubc.ca}
}
\maketitle
\begin{center}
{\normalfont
\vspace{-2em}
$^{1}$ McMaster University \hspace{0.8em}
$^{2}$ University of British Columbia \\
$^{3}$ Vector Institute \hspace{0.8em}
$^{4}$ Viggle AI \hspace{0.8em}
$^{5}$ Canada CIFAR AI Chair
\vspace{1em}
}
\end{center}

\begin{abstract}
   Predicting future motion is crucial in video understanding and controllable video generation. Dense point trajectories are a compact, expressive motion representation, but modeling their future evolution from observed video remains challenging. We propose a framework that predicts future trajectories and visibility from past trajectories and video context. Our method has three components: (1) \encodingname{}, which reduces location-dependent bias by representing each point as an offset from its pixel-center anchor; (2) \vaename{}, which learns a compact spatiotemporal latent space for dense trajectories with masked reconstruction and a spatiotemporal consistency regularizer; and (3) \modelname{}, which generates future trajectories in latent space via flow matching, with boundary cues and on-policy $K$-step fine-tuning for stable sampling. We also introduce \benchmarkname{}, a unified benchmark spanning real and synthetic videos with a standardized setup aligned with video-generation benchmarks. Compared with state-of-the-art methods, our approach extends the prediction horizon from 24 to 81 frames while improving motion realism and stability across datasets. The predicted trajectories directly support downstream video generation and editing. We released code, model checkpoints, and datasets at \url{https://trajloom.github.io/}.
\end{abstract}

\section{Introduction}

Motion is central to video and carries information beyond static appearance~\cite{simonyan2014two}.
Recent video generation and editing systems rely on motion cues---including camera control, optical flow, and trajectory guidance---to shape temporal dynamics~\cite{geng2025motion,burgert2025go,wang2024motionctrl,chu2025wanmove,deng2024dragvideo}.
Point trajectories are a flexible motion representation. Modern trackers can recover dense trajectories with long-range correspondences and occlusion patterns~\cite{harley2025alltracker,karaev2025cotracker3,wang2023tracking,doersch2022tap}.
This motivates a key question: given trajectories in a fixed history window, how can we predict their future positions and visibility over a future horizon?

Trajectory forecasting methods model future motion directly in trajectory space~\cite{walker2016uncertain,wen2023anypoint,bharadhwaj2024track2act,yang2025tra}.
However, future motion is inherently uncertain and multimodal, making deterministic prediction insufficient.
A recent method, \emph{What Happens Next?} (WHN)~\cite{boduljak2026what}, formulates trajectory anticipation as a generative task and is primarily conditioned on appearance cues in a given image and possibly text prompts.
However, appearance-only conditioning overlooks explicit motion history. 
Observed trajectories already encode current dynamics and strongly constrain plausible futures.
This motivates future-trajectory generation conditioned on trajectory and video history. 
The main challenges are preserving temporal stability and local coherence across forecast windows in diverse real-world videos.



In contrast to image-conditioned trajectory generators, we forecast from observed trajectory and video history. 
This conditioning captures ongoing dynamics and differs from WHN-style appearance-driven generation, which mainly depends on image content~\cite{boduljak2026what}. 
A central design question is how to represent dense trajectories for learning. 
Most methods use \emph{absolute} image coordinates~\cite{boduljak2026what}, which couple motion with global position and induce location-dependent statistics. 
We instead propose \encodingname{}, which represents each trajectory as a displacement from a fixed pixel-center anchor. 
Absolute coordinates are recovered by adding anchors back. 
This offset-based parameterization emphasizes motion rather than location and provides a stable foundation for latent generative modeling.

Even with \encodingname{}, forecasting dense trajectory fields remains high-dimensional. 
We first learn \vaename{}, a variational autoencoder (VAE)~\cite{Kingma2013AutoEncodingVB} that maps trajectory segments to compact spatiotemporal tokens and reconstructs dense tracks. 
To preserve motion structure, \vaename{} applies a spatiotemporal consistency regularizer that aligns velocities with local neighbors. 
We then generate future motion in this latent space using \modelname{}, a rectified-flow model conditioned on observed trajectories and video that predicts the full future window~\cite{lipmanflow,liu2023flow}. 
Lightweight boundary cues enforce continuity with observed history. 
Because training uses constructed interpolation states whereas inference queries self-visited ODE states, we further use on-policy $K$-step fine-tuning to reduce this mismatch.
For evaluation, we introduce \benchmarkname{}, a unified benchmark spanning real and synthetic videos with standardized setups (\eg, resolution and horizon) as common video generation benchmarks~\cite{doersch2022tap,vecerik2024robotap,greff2021kubric,Li_2025_ICCV}. Compared with WHN~\cite{boduljak2026what}, our method improves motion realism, temporal consistency, and stability in both quantitative and qualitative evaluations. We also show that the predicted trajectories effectively guide motion-controlled video generation and editing~\cite{wan2025wan,chu2025wanmove}.

We summarize our main contributions as follows.
\begin{enumerate}
    \item \textbf{Trajectory encoding:} \encodingname{}, which represents each point as an offset from a fixed grid anchor to reduce location-dependent bias in dense trajectory prediction.
    \item \textbf{Latent trajectory generation:} A generative framework that combines (i) \vaename{}, a VAE with masked reconstruction and spatiotemporal regularization for compact, structured trajectory latents, and (ii) \modelname{}, a rectified-flow generator conditioned on observed trajectories and video, with boundary cues and on-policy $K$-step fine-tuning for stable sampling over extended forecast windows.
    \item \textbf{Benchmark and results:} \benchmarkname{}, a unified benchmark for dense trajectory forecasting in natural videos. Our approach achieves state-of-the-art performance and provides a strong foundation for downstream applications such as motion-controlled video generation and editing.
\end{enumerate}

\section{Related Works}
\begin{figure}[ht]
    \centering
    \includegraphics[width=\linewidth]{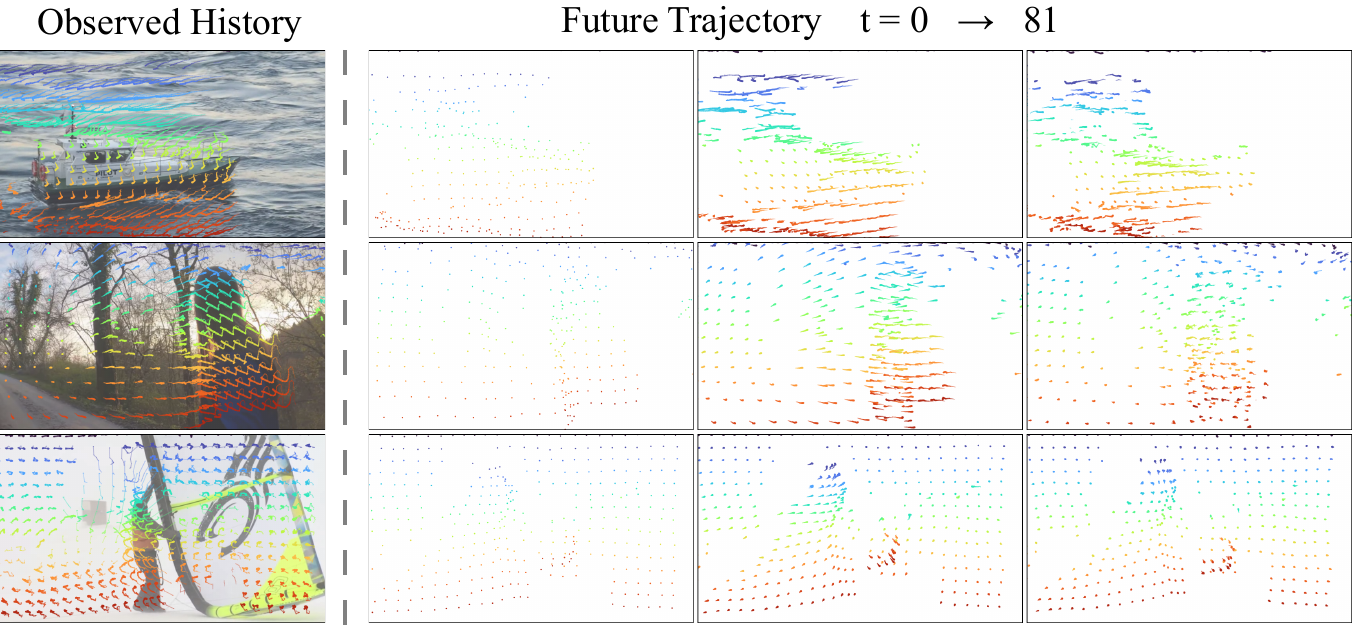}
\caption{For each sequence, the model observes an 81-frame history (left) and predicts future trajectories for the next 81 frames (right). Predicted trajectories are shown at three times: early, middle, and final. Colors show the spatial order of query points.}
    \label{fig:trajectory-view-1}
\end{figure}

\noindent \textbf{Trajectories for motion anticipation.}
Modern tracking-any-point methods track long-range point trajectories (with visibility/occlusion) in unconstrained videos, enabling dense correspondence under large motion and occlusions~\cite{doersch2022tap,doersch2023tapir,doersch2024bootstap,karaev2025cotracker3,wang2023tracking,harley2025alltracker,harley2022particle}.
Recent datasets and training pipelines further scale tracking quality and diversity, e.g., PointOdyssey for long synthetic sequences and BootsTAP for leveraging unlabeled real video~\cite{zheng2023pointodyssey,doersch2024bootstap}.
Given an observed history window of tracks and visibility, \emph{trajectory prediction} forecasts future positions directly in trajectory space. It has been used for forecasting, planning, and imitation in robotics and action reasoning~\cite{walker2016uncertain,vecerik2024robotap,wen2023anypoint,bharadhwaj2024track2act,yang2025tra}.
Most approaches remain regression-based and can average over multiple plausible futures, becoming conservative and accumulating drift over long horizons.
This motivates formulating future motion as generative. \emph{What Happens Next?} (WHN) samples dense future trajectories from appearance cues like image or text, instead of predicting a single deterministic continuation~\cite{boduljak2026what}.
Our work follows this generative direction but conditions on the observed motion history, leveraging constraints already present in tracked trajectories.


\noindent \textbf{Motion-guided generation and editing.}
Controllable video generation often incorporates explicit motion controls such as optical flow, camera trajectories, or point tracks to guide temporal dynamics in diffusion-based models~\cite{wang2024motionctrl,burgert2025go,geng2025motion,wang2023videocomposer}.
Trajectory-conditioned methods use sparse or dense tracks as a low-level interface for directing object motion, as exemplified by DragNUWA, MagicMotion, Tora, and SG-I2V~\cite{yin2023dragnuwa,Li_2025_ICCV,zhang2025tora,namekata2025sgiv}.
Wan-Move is particularly relevant to our applications. Built on the Wan image-to-video backbone, it employs latent trajectory guidance that propagates information along dense point trajectories, enabling direct point-level motion control~\cite{chu2025wanmove,wan2025wan}.
Interactive editing similarly uses sparse point constraints to manipulate deformation and motion. These range from DragGAN and DragDiffusion to video drag methods such as DragVideo~\cite{pan2023drag,shi2024dragdiffusion,zhang2025gooddrag,deng2024dragvideo}.
Our future-trajectory generator complements these controllers. We adopt Wan-Move because it consumes dense trajectories \emph{directly}, allowing our predicted tracks to integrate without additional motion representations~\cite{chu2025wanmove}.

\section{TrajLoom: Dense Future Trajectory Generation}
\label{sec:methods}

\begin{figure}[ht]
    \centering
    \includegraphics[width=\linewidth]{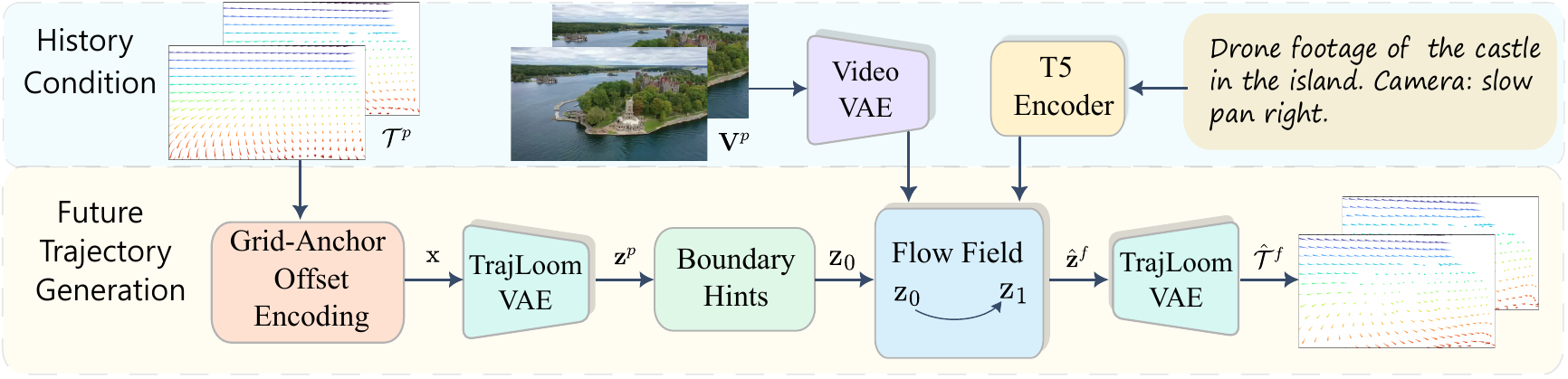}
\caption{\textbf{Overview of our pipeline.} Given observed trajectories $\mathcal{T}^{p}$, we rasterize and encode them with \encodingname{} into a dense offset field, then compress with \vaename{} into history latents $\mathbf{z}^{p}$. Conditioned on $\mathbf{z}^{p}$ and video features, \modelname{} generates future latents via rectified-flow integration with boundary hints, which are decoded by \vaename{} into future trajectories $\hat{\mathcal{T}}^{f}$.}
    \label{fig:pipeline}
\end{figure}

\noindent We study future-motion generation from observed history in a video clip.
A video is denoted as $\mathbf{V}\in\mathbb{R}^{T\times H\times W\times 3}$, where $T$ is the total number of frames, and each frame has a spatial resolution of $H\times W$.
Motion is represented as a set of $N$ trajectories, $\mathcal{T}=\{\tau_n\}_{n=1}^{N}$, where each trajectory $\tau_n=\{(x_{n,t},y_{n,t})\}_{t=0}^{T-1}$ tracks one reference 2D point through time.
At each frame $t$, the location of the point is $(x_{n,t},y_{n,t})$, accompanied by a visibility indicator $v_{n,t}\in\{0,1\}$.
For clarity, we split each clip into a past history window of length $T_p$ and a future window of length $T_f$, where $T_p+T_f=T$.
Thus, $\mathbf{V}$ is divided as $\mathbf{V}=(\mathbf{V}^{p},\mathbf{V}^{f})$. Each trajectory $\tau_n$ is similarly partitioned into a history segment, $\tau_n^{p}=\{(x_{n,t},y_{n,t})\}_{t=0}^{T_p-1}$, and a future segment, $\tau_n^{f}=\{(x_{n,t},y_{n,t})\}_{t=T_p}^{T-1}$. Visibility indicators are split in the same manner.
Given the observed history trajectories $\mathcal{T}^{p}=\{\tau_n^{p}\}_{n=1}^{N}$, their corresponding visibility indicators, the history video clip $\mathbf{V}^{p}$, and a text caption, our goal is to generate the corresponding future trajectories $\hat{\mathcal{T}}^{f}$.


Our pipeline has three stages: \encodingname{} densifies sparse trajectories into grid-anchored offsets (Section~\ref{sec:encoding}); \vaename{} compresses dense fields into compact spatiotemporal latents with masked reconstruction and spatiotemporal regularization (Section~\ref{sec:vae}); and \modelname{} jointly predicts future latents via a history-conditioned rectified flow, then decodes them into trajectories (Section~\ref{sec:generator}). To reduce train-test mismatch from ODE integration~\cite{chen2018neural}, we further apply on-policy $K$-step fine-tuning.

\subsection{\encodingname}
\label{sec:encoding}
\begin{figure}[tb]
  \centering
  \begin{subfigure}{0.59\linewidth}
    \includegraphics[width=\linewidth]{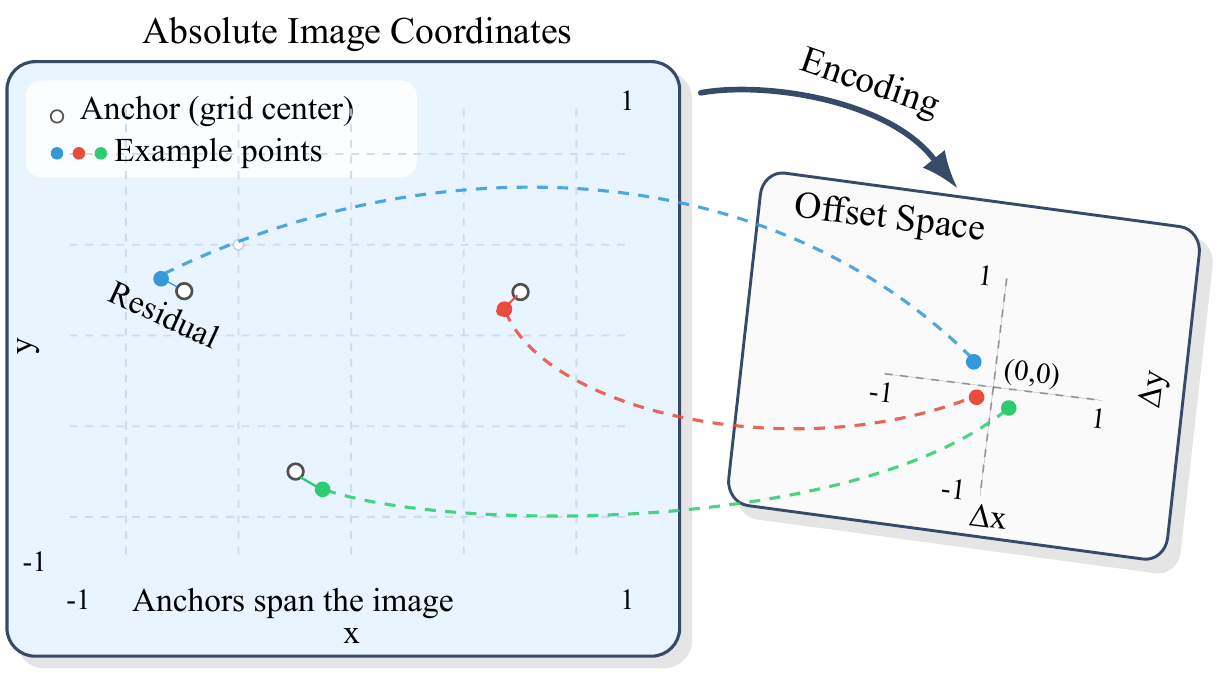}
    \caption{\textbf{Grid-Anchored Offset Encoding.} We encode trajectories as displacements from pixel-center anchors: instead of absolute coordinates (left), each point is given by its offset $(\Delta x,\Delta y)$ from its local anchor (right), producing a zero-centered, location-consistent representation. }
    \label{fig:offset}
  \end{subfigure}
  \hfill
  \begin{subfigure}{0.39\linewidth}
  \centering
    \includegraphics[width=0.75\linewidth]{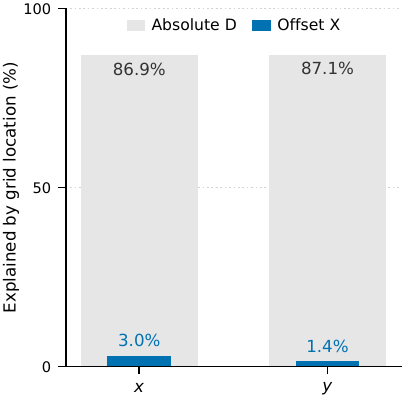}
\caption{Variance explained by grid location for absolute and offset coordinates. We measure coordinate variance attributable to grid position using time-averaged coordinates at each grid cell.}
    \label{fig:offset-insight-metric}
  \end{subfigure}
\caption{\encodingname{} converts absolute trajectories into offset space, reducing the bias of absolute coordinates.}
  \label{fig:offset_two_figs}
\end{figure}

Starting from trajectories $\mathcal{T}=\{\tau_n\}_{n=1}^{N}$, \encodingname{} constructs a dense trajectory representation on the video grid.
\encodingname{} represents each pixel by its displacement from a local pixel-center anchor, rather than by absolute coordinates. This yields offsets that are consistent and comparable across grid locations.

Concretely, trajectories are extracted from a stride-$s$ grid.
Let $H_c=H/s$ and $W_c=W/s$ so that $N=H_cW_c$.
Rasterization produces (i) a dense absolute coordinate field $\mathbf{D}\in\mathbb{R}^{T\times H\times W\times 2}$ and (ii) a dense visibility mask $\mathbf{M}\in\{0,1\}^{T\times H\times W}$.
For a pixel location $p=(h,w)$, the corresponding coarse-grid trajectory index is $\pi(h,w)=\Bigl\lfloor {h}/{s}\Bigr\rfloor W_c+\Bigl\lfloor {w}/{s}\Bigr\rfloor+1$
and the dense fields are defined by
$\mathbf{D}(t,h,w)=\bigl(x_{\pi(h,w),t},y_{\pi(h,w),t}\bigr)$ and the mask is $\mathbf{M}(t,h,w)=v_{\pi(h,w),t}$.
By construction, $\mathbf{D}$ and $\mathbf{M}$ are piecewise constant within each $s\times s$ stride cell.
All coordinates are represented in a normalized image coordinate system.

For each pixel $p$ at location $(h, w)$ in the dense field, we define its normalized pixel-center anchor $\mathbf{G}(p)\in\mathbb{R}^2$, where $\mathbf{G}(p)=\left[\,2\frac{w+\frac{1}{2}}{W}-1,\;2\frac{h+\frac{1}{2}}{H}-1\,\right]^\top$.
The offset-encoded trajectory field is then $\mathbf{X}(t,p)=\mathbf{D}(t,p)-\mathbf{G}(p)$.
From now on, the offset field $\mathbf{X}$, together with the visibility mask $\mathbf{M}$, serves as the trajectory representation.
Absolute coordinates can be recovered from this representation.
We validate \encodingname{} by comparing coordinate variance under absolute and relative representations.
With absolute coordinates $\mathbf{D}$, trajectory variance is dominated by grid location: points from different grid cells are centered at different image positions, so the overall variance is large even when local motion is similar.
To quantify this effect, we compute the fraction of coordinate variance explained by grid location, using a visibility-weighted time-averaged coordinate at each grid position as the location baseline.
Figure~\ref{fig:offset-insight-metric} shows that this explained variance is high for absolute coordinates but much lower for relative offsets.
Using offsets $\mathbf{X}=\mathbf{D}-\mathbf{G}$ removes most location-driven variance and yields a more uniform representation focused on local displacement.
More details can be found in Appendix~\ref{app:offset_insight}. 

\subsection{\vaename}
\label{sec:vae}

Modeling future motion directly in dense trajectory-field space is high-dimensional.
To obtain a compact representation for generative modeling, we learn a variational autoencoder (VAE) in the latent space.
\vaename{} is trained on temporal segments $\mathbf{x}$ from the offset-encoded trajectory field $\mathbf{X}$ (Section~\ref{sec:encoding}) and the corresponding visibility mask $\mathbf{m}$.
Given a segment $\mathbf{x}$, the encoder defines an approximate posterior $q_\phi(\mathbf{z}\mid\mathbf{x})$, and the decoder reconstructs it as $\hat{\mathbf{x}}=\psi(\mathbf{z})$.

A masked pointwise reconstruction loss encourages $\hat{\mathbf{x}}$ to match $\mathbf{x}$ at visible locations, but it does not directly model temporal evolution or local relative motion.
As a result, reconstructions with reconstruction error alone still show temporal jitter or local spatial inconsistency (see Appendix~\ref{app:vae_toy}).
To enforce temporal smoothness and local coherence, we propose a \emph{spatiotemporal consistency regularizer} that matches (i) temporal velocities and (ii) multiscale spatial neighbor relations between the target segment $\mathbf{x}$ and reconstruction $\hat{\mathbf{x}}$.


\subsubsection{Spatiotemporal consistency regularizer.}
The regularizer combines a temporal velocity term and a multiscale spatial neighbor term.
Let $\Omega$ denote the set of spacetime indices $(t,p)$ within a segment window.
The trajectory value at $(t,p)$ is $\mathbf{x}(t,p)\in\mathbb{R}^2$, and the corresponding visibility is $\mathbf{m}(t,p)\in\{0,1\}$.
All consistency terms are computed only on valid, visible pairs and are normalized by the number of such pairs so that the loss scale does not depend on how many points are visible.

We discourage the frame-to-frame jittering by the following loss, 
\begin{equation}
  L_{\mathrm{temporal}}
  =
  \frac{1}{\sum_{(t,p)\in\Omega}\mathbf{m}_{\mathrm{pair}}(t,p)}
  \sum_{(t,p)\in\Omega}\mathbf{m}_{\mathrm{pair}}(t,p)\,
  \left\|\Delta_i \hat{\mathbf{x}}(t,p)-\Delta_i \mathbf{x}(t,p)\right\|_1,
\end{equation}
where $\Delta_i \mathbf{x}(t,p)=\mathbf{x}(t,p)-\mathbf{x}(t-1,p)$, $\Delta_i \hat{\mathbf{x}}(t,p)=\hat{\mathbf{x}}(t,p)-\hat{\mathbf{x}}(t-1,p)$, and $\mathbf{m}_{\mathrm{pair}}(t,p)=\mathbf{m}(t,p)\mathbf{m}(t-1,p)$.
Bascially, it matches the temporal consistency between the reconstruction and the ground truth for locations that are visible.

To preserve spatial consistency, we additionally match relative motion among neighboring locations.
Let $\mathcal{S}$ be a set of horizontal/vertical offsets at multi-hop distances.
For each neighboring location $\delta\in\mathcal{S}$, we define $\Delta_\delta \mathbf{x}(t,p)=\mathbf{x}(t,p+\delta)-\mathbf{x}(t,p)$ and $\Delta_\delta \hat{\mathbf{x}}(t,p)=\hat{\mathbf{x}}(t,p+\delta)-\hat{\mathbf{x}}(t,p)$, and introduce the following loss,
\begin{equation}
  L_{\mathrm{spatial}}
  =
  \frac{1}{\sum_{\delta\in\mathcal{S}}\alpha_\delta}
  \sum_{\delta\in\mathcal{S}}
  \alpha_\delta\;
  \frac{
    \sum_{(t,p)\in\Omega}\mathbf{m}_\delta(t,p)\,
    \left\|\Delta_\delta \hat{\mathbf{x}}(t,p)-\Delta_\delta \mathbf{x}(t,p)\right\|_1
  }{
    \sum_{(t,p)\in\Omega}\mathbf{m}_\delta(t,p)
  }.
\end{equation}
The loss is only activated when both neighboring locations are visible since $\mathbf{m}_\delta(t,p)=\mathbf{m}(t,p)\mathbf{m}(i,p+\delta)$.
Each neighbor is weighted by $\alpha_\delta$ and then normalized by the sum over the neighborhood. The set $\mathcal{S}=\{1,2,4\}$ determines the hop distances used, with the corresponding $\Delta_\delta$ values of 1, 0.5, and 0.25. We scale down the $\alpha_\delta$ by the neighborhood distance; the larger the distance $\delta$, the smaller $\alpha_\delta$. This makes the spatial loss $L_{\mathrm{spatial}}$ focus more on local motion, since the global motion is captured by the reconstruction loss.

Therefore, the full spatiotemporal consistency regularizer is,
\begin{equation}
  L_{\mathrm{st}}
  =
  \lambda_{\mathrm{temporal}}\,L_{\mathrm{temporal}}
  +
  \lambda_{\mathrm{spatial}}\,L_{\mathrm{spatial}},
\end{equation}
where $\lambda_{\mathrm{temporal}}$ and $\lambda_{\mathrm{spatial}}$ are weighting coefficients.

Appendix~\ref{app:vae_toy} and Figure~\ref{fig:vae-toy-jitter} provide a toy example showing why pointwise reconstruction alone is insufficient and how the consistency regularizer separates smooth from jittery solutions.

\subsubsection{Training objective.}

The reconstruction loss of our VAE is as follows, 
\begin{equation}
  L_{\mathrm{rec}}
  =
  \sum_{(t,p)\in\Omega}
  w(t,p)\,
  \rho\bigl(\hat{\mathbf{x}}(t,p)-\mathbf{x}(t,p)\bigr),
\end{equation}
where the normalized mask $w(t,p)=\mathbf{m}(t,p)\big/\sum_{(j,q)\in\Omega}\mathbf{m}(j,q)$ ensures that we only consider visible locations.
$\rho$ is the Huber loss~\cite{huber1992robust}.

We train \vaename{} by minimizing reconstruction error, the KL divergence, and the spatiotemporal consistency regularizer,
\begin{equation}
  L_{\mathrm{vae}}
  =
  \mathbb{E}_{q_\phi(\mathbf{z}\mid \mathbf{x})}\!\left[L_{\mathrm{rec}}+L_{\mathrm{st}}\right]
  +\beta\,
  D_{\mathrm{KL}}\!\left(q_\phi(\mathbf{z}\mid \mathbf{x})\,\|\,\mathcal{N}(\mathbf{0},\mathbf{I})\right),
\end{equation}
where $\beta$ is the weighting of the KL term. In practice, we set $\beta=5{\times}10^{-5}$, $\lambda_{\mathrm{temporal}}$ as 0.1, and $\lambda_{\mathrm{spatial}}$ as 0.2 for the spatiotemporal consistency regularizer.

\subsection{\modelname}
\label{sec:generator}

We generate future motion in the latent space learned by \vaename{}.
Given a history segment $\mathbf{x}^{p}$ and a future segment $\mathbf{x}^{f}$ (both from the offset field $\mathbf{X}$), we obtain their latent representations with the frozen VAE encoder.
We use the posterior mean as a deterministic encoding:
\begin{equation}
  \mathbf{z}^{p}=\mathbb{E}_{q_\phi(\mathbf{z}\mid \mathbf{x}^{p})}[\mathbf{z}],
  \qquad
  \mathbf{z}^{f}=\mathbb{E}_{q_\phi(\mathbf{z}\mid \mathbf{x}^{f})}[\mathbf{z}].
\end{equation}
\modelname{} models the conditional distribution of future latents given observed history and predicts the full future window jointly.

To keep predictions consistent with observed motion, we summarize all conditioning signals as $\mathbf{c}$.
In our setting, $\mathbf{c}$ includes history trajectory latents $\mathbf{z}^{p}$, history visibility, and history-video features.
The generator is a latent flow matching model, parameterized by a conditional velocity field $v_\theta(\mathbf{z}_t,t,\mathbf{c})$.

\subsubsection{Boundary hints.}
\label{sec:generator_hints}
\begin{wrapfigure}{r}{0.5\textwidth} 
    \centering
    \includegraphics[width=1\linewidth]{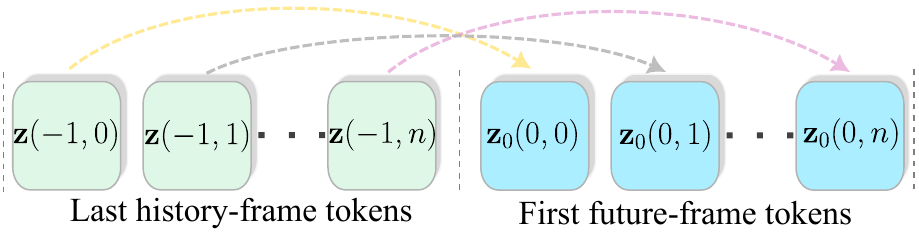}
\caption{We initialize $\mathbf{z}_0$ from scaled Gaussian noise, then add each last history token $\mathbf{z}(-1,n)$ to the first future token $\mathbf{z}_0(0,n)$.}
    \label{fig:boundary}
\end{wrapfigure}
Because we generate the entire future window jointly, we provide explicit boundary information so the model can align future predictions with the observed past.
We use two lightweight mechanisms: (i) a boundary-anchored initialization of $\mathbf{z}_0$, and (ii) token-aligned fusion of history latents into the query stream.


Let $\Lambda=\{(k,n)\}$ index latent tokens, where $k$ denotes a latent time index and $n$ denotes a spatial token index, and let $\mathbf{z}(k,n)\in\mathbb{R}^{C}$ denote a token. Denoting by $\mathbf{z}(-1,n)$ the latent at the last history time step, we initialize the source state by repeating this boundary latent across the future horizon and adding Gaussian noise:
\begin{equation}
  \mathbf{z}_0(k,n)=\mathbf{z}(-1,n)+\sigma_0\,\boldsymbol{\eta}(k,n),
  \qquad
  \boldsymbol{\eta}\sim\mathcal{N}(\mathbf{0},\mathbf{I}),
\end{equation}
where $\sigma_0$ controls the noise scale. In practice, we apply this anchoring at $k=0$.

Beyond conditioning through $\mathbf{c}$, we inject history latents $\mathbf{z}^{p}$ into the velocity network through a small token-aligned fusion module, providing a direct boundary cue. More details are in Appendix~\ref{app:boundary} and the ablation study in Appendix~\ref{app:generator_ablation}.

\subsubsection{Flow matching.}
\label{sec:generator_fm}

To model a distribution over future latents without autoregressive rollout, we adopt rectified flow~\cite{lipmanflow,liu2023flow} and learn a conditional latent velocity field.
Denote the future target as $\mathbf{z}_1=\mathbf{z}^{f}$, and let $\mathbf{z}_0$ be the history-conditioned source state.
For flow time $t\in[0,1]$, an intermediate state is
\begin{equation}
  \mathbf{z}_t=(1-t)\,\mathbf{z}_0+t\,\mathbf{z}_1+\sigma\,\boldsymbol{\epsilon},
  \qquad
  \boldsymbol{\epsilon}\sim\mathcal{N}(\mathbf{0},\mathbf{I}),
\end{equation}
and the model predicts a conditional velocity field $v_\theta(\mathbf{z}_t,t,\mathbf{c})$.
Under linear interpolation, the target velocity is $u_t=\mathbf{z}_1-\mathbf{z}_0$, and training matches $v_\theta$ to $u_t$.

To emphasize visible future regions, we obtain a token-level weight $\mathbf{m}^{\mathrm{tok}}(k,n)\in[0,1]$ by pooling the future visibility mask onto the VAE token grid.
We define normalized token weights as $w_\Lambda(k,n) = {\mathbf{m}^{\mathrm{tok}}(k,n)}/{\sum_{(j,q)\in\Lambda}\mathbf{m}^{\mathrm{tok}}(j,q)}$.
The resulting visibility-weighted flow-matching loss is
\begin{equation}
  L_{\mathrm{fm}}
  =
  \mathbb{E}_{t,\boldsymbol{\epsilon}}
  \left[
    \frac{1}{C}
    \sum_{(k,n)\in\Lambda}
    w_\Lambda(k,n)\,
    \left\|
      v_\theta(\mathbf{z}_t,t,\mathbf{c})(k,n)-u_t(k,n)
    \right\|_2^2
  \right],
\end{equation}
where $C$ denotes the number of latent channels in $\mathbf{z}$.

\subsubsection{On-policy fine-tuning.}
\label{sec:generator_kstep}

Flow matching trains $v_\theta$ using interpolated states $\mathbf{z}_t$, while sampling evaluates $v_\theta$ on states produced by integrating the learned ODE.
This mismatch can cause drift because the model is queried off the training path.
We therefore apply an on-policy $K$-step rollout loss to fine-tune $v_\theta$ on its own visited states.

Let $\varepsilon=t_0<t_1<\cdots<t_K=1-\varepsilon$ be an increasing time grid and set $\tilde{\mathbf{z}}_0=\mathbf{z}_0$.
A detached forward-Euler rollout generates visited states
\begin{equation}
  \tilde{\mathbf{z}}_{i+1}
  =
  \tilde{\mathbf{z}}_{i}
  +
  (t_{i+1}-t_i)\,
  \mathrm{sg}\!\left[v_\theta(\tilde{\mathbf{z}}_i,t_i,\mathbf{c})\right],
  \qquad i=0,\dots,K-1.
  \label{eq:kstep_rollout}
\end{equation}
Denote $\mathbf{v}_i = v_\theta(\tilde{\mathbf{z}}_i,t_i,\mathbf{c})$.
Endpoint-consistent velocity targets are
\begin{equation}
  \mathbf{v}^{(1)}_i = \frac{\mathbf{z}_1-\tilde{\mathbf{z}}_i}{1-t_i},
  \qquad
  \mathbf{v}^{(0)}_i = \frac{\tilde{\mathbf{z}}_i-\mathbf{z}_0}{t_i}.
  \label{eq:kstep_targets}
\end{equation}
The on-policy rollout loss is defined as,
\begin{equation}
  L_{\text{k-step}}
  =
  \frac{1}{K}\sum_{i=0}^{K-1}
  \Bigl(
    w_{1}\,\|\mathbf{v}_i-\mathbf{v}^{(1)}_i\|_{w_\Lambda}^2
    +
    w_{0}\,\|\mathbf{v}_i-\mathbf{v}^{(0)}_i\|_{w_\Lambda}^2
  \Bigr),
  \label{eq:kstep_loss}
\end{equation}
where $\| f \|_{w_\Lambda}^2 = \frac{1}{C} \sum_{(k,n)\in\Lambda} w_\Lambda(k,n)\ \left\| f(k,n) \right\|_2^2$.
We further introduce a simple endpoint-consistency term to stabilize implied endpoints along the rollout,
\begin{equation}
  L_{\mathrm{cons}}
  =
  \frac{1}{K-1}\sum_{i=1}^{K-1}
  \Bigl(
    \|\hat{\mathbf{z}}^{(1)}_i-\mathrm{sg}[\hat{\mathbf{z}}^{(1)}_{i-1}]\|_{w_\Lambda}^2
    +
    \|\hat{\mathbf{z}}^{(0)}_i-\mathrm{sg}[\hat{\mathbf{z}}^{(0)}_{i-1}]\|_{w_\Lambda}^2
  \Bigr),
  \label{eq:kstep_cons}
\end{equation}
where $\hat{\mathbf{z}}^{(1)}_i=\tilde{\mathbf{z}}_i+(1-t_i)\mathbf{v}_i$ and $\hat{\mathbf{z}}^{(0)}_i=\tilde{\mathbf{z}}_i-t_i\mathbf{v}_i$.
The final loss is,
\begin{equation}
  L
  =
  L_{\mathrm{fm}}
  +
  \lambda_{\text{k-step}}
  \Bigl(L_{\text{k-step}}+\gamma\,L_{\mathrm{cons}}\Bigr).
  \label{eq:gen_final_obj}
\end{equation}
In practice, we apply this loss on a small sub-batch to limit overhead, and use small $\lambda_{\text{k-step}}$ and $\gamma$ to stabilize training. More details are in Appendix~\ref{app:onpolicy}.

\subsubsection{Sampling.}

At inference, we obtain future latents by integrating the learned rectified-flow ODE from the history-conditioned source state.
We take the final state as the generated future latent $\hat{\mathbf{z}}^{f}=\mathbf{z}(1)$.
Finally, $\hat{\mathbf{z}}^{f}$ is decoded using the frozen \vaename{} decoder to obtain future dense trajectories.

\section{Experiments}
We evaluate both components of our framework: \vaename{} for trajectory reconstruction and \modelname{} for future trajectory generation.

\subsection{Implementation Details}

\subsubsection{Baseline.}
We compare against WHN (L), the largest variant of WHN~\cite{boduljak2026what}, a state-of-the-art image-conditioned dense trajectory generator.

\subsubsection{Trajectory extraction.}
\label{sec:traj_extract}
Our framework is trained and evaluated on dense trajectory fields $\mathbf{X}$ and visibility masks $\mathbf{M}$ (Section~\ref{sec:methods}). Each video is converted to $(\mathbf{X},\mathbf{M})$ via dense point tracking followed by rasterization.
For all datasets, we extract dense long-range trajectories $\mathcal{T}$ with AllTracker~\cite{harley2025alltracker}, using the first frame as reference and a stride-32 grid, for both training and evaluation.

\subsubsection{Training dataset.}
Training uses MagicData~\cite{Li_2025_ICCV}, a motion-focused text--video dataset with about 23k video--caption pairs.
We apply standard filtering by aspect ratio, resolution, and clip length for consistency.
Following WAN~\cite{wan2025wan}, videos are processed at 480p, and clips shorter than 162 frames are removed.
After filtering, 16k videos remain; we use the first 162 frames of each video to match our forecasting window.
We split these samples 90\%/10\% for training and validation.


\subsubsection{Benchmark.}
Evaluation uses \benchmarkname{}, introduced in this work. It includes real and synthetic videos aggregated from existing datasets, covers diverse dense-forecasting scenarios, and uses MagicData validation. We then apply a unified resolution, temporal length, and preprocessing pipeline for fair comparison.

\noindent (i) \textit{Real-world sources.}
TAP-Vid evaluation sources~\cite{doersch2022tap} are reconstructed from raw videos.
Specifically, TAP-Vid-Kinetics is constructed from YouTube IDs and temporal segments from the Kinetics-700 validation set~\cite{DBLP:journals/corr/KayCSZHVVGBNSZ17}, and RoboTAP~\cite{vecerik2024robotap} is provided in the same point-track annotation format as other TAP-Vid-style datasets.
Instead of using TAP-Vid resized videos, we re-extract all videos at the target resolution and convert them into fixed-length temporal windows to match the forecasting horizon.

\noindent (ii) \textit{Synthetic sources.} 
In its original work, WHN (L)~\cite{boduljak2026what} is trained and evaluated on a Kubric variant (MOVi-A)~\cite{greff2021kubric}, enabling direct comparison on this dataset.
Kubric is also a primary synthetic source in TAP-Vid~\cite{doersch2022tap}.
For comparability, we follow the same MOVi-A configuration as WHN and re-render longer videos when needed.

\subsubsection{Model and training.}
Following WHN~\cite{boduljak2026what}, we use a DiT backbone~\cite{peebles2023scalable}, specifically Latte~\cite{ma2025latte}, for both \vaename{} and \modelname{}.
\vaename{} uses 16 blocks, 8 attention heads, 512 hidden dimensions, and 16 latent channels, plus a temporal convolution layer for temporal downsampling.
\modelname{} follows the WHN (L) scale: 16 blocks, 12 heads, and 768 hidden dimensions.
We train with AdamW~\cite{DBLP:journals/corr/KingmaB14, loshchilov2018decoupled} at learning rate $6\times10^{-5}$, and use $1\times10^{-5}$ for on-policy fine-tuning.
Additional architecture and hyperparameter details are provided in Appendix~\ref{app:train-model}.

\subsubsection{Evaluation Metrics.}
\label{sec:metrics}
We report evaluation metrics for (i) reconstruction fidelity of \vaename{} and (ii) motion quality of \modelname{}.
For the generator, we use FVMD~\cite{liu2024fr} to evaluate motion realism and temporal consistency, and two reference-free diagnostics (FlowSmoothTV and DivCurlEnergy) to assess trajectory quality.
For the VAE, we report visibility-masked endpoint error (VEPE) as the reconstruction metric.


\noindent (i) \textit{Fr\'echet Video Motion Distance (FVMD).}
FVMD~\cite{liu2024fr} uses Fr\'echet distance to compare motion-feature distributions derived from point trajectories.
Unless noted otherwise, we use the official setting (clip length 16, stride 1) and restrict feature extraction to visible points via the trajectory visibility mask.
Standard FVMD uses short 16-frame clips to capture fine-grained motion.
For long-horizon forecasting (81 frames), however, errors often accumulate late and can be underrepresented by short clips. We therefore also report FVMD-Long, computed on the full 81-frame future window as a single clip.

\noindent (ii) \textit{FlowSmoothTV (FlowTV)} captures \emph{spatial tearing}, where neighboring grid cells move inconsistently and produce large spatial flow gradients.
We measure it as the total variation of per-frame flow on the grid, following TV-regularized optical flow~\cite{zach2007duality}.
Let $\hat{\mathbf{p}}_{t,i,j}\in\mathbb{R}^2$ be the predicted location at time $t$ and define the flow
$\mathbf{f}_{t,i,j}=(u_{t,i,j},v_{t,i,j})=\hat{\mathbf{p}}_{t,i,j}-\hat{\mathbf{p}}_{t-1,i,j}$ for $t\ge 1$.
With grid spacing $s$ (pixels), we use forward differences
$\Delta_x u_{t,i,j}=(u_{t,i,j+1}-u_{t,i,j})/s$ and
$\Delta_y u_{t,i,j}=(u_{t,i+1,j}-u_{t,i,j})/s$ (similarly for $v$),
computed only at valid (visibility is true) neighbor pairs.
For each time step, we define
$\mathrm{TV}_x(t)=\mathbb{E}^{x}_{i,j}\!\left[|\Delta_x u_{t,i,j}|+|\Delta_x v_{t,i,j}|\right]$ and
$\mathrm{TV}_y(t)=\mathbb{E}^{y}_{i,j}\!\left[|\Delta_y u_{t,i,j}|+|\Delta_y v_{t,i,j}|\right]$,
where $\mathbb{E}^{x}$ and $\mathbb{E}^{y}$ average over valid horizontal and vertical pairs, respectively.
We then average it over time:
\begin{equation}
  \mathrm{FlowTV}
  =
  \frac{1}{T-1}\sum_{t=1}^{T-1}\bigl(\mathrm{TV}_x(t)+\mathrm{TV}_y(t)\bigr).
\end{equation}
Lower FlowTV means fewer spatial discontinuities.

\noindent (iii) \textit{DivCurlEnergy (DivCurlE).}
DivCurlEnergy~\cite{corpetti2006fluid} captures \emph{locally unstable deformation}, where predicted flow shows abrupt local expansion, contraction, or rotation.
It measures large divergence (expansion/contraction) and curl (local rotation) in the predicted flow field.
Using the same flow $\mathbf{f}_{t,i,j}=(u_{t,i,j},v_{t,i,j})$, we form forward-difference divergence and curl on valid grid cells as
$\mathrm{div}(\mathbf{f}_t)=(\Delta_x u+\Delta_y v)/s$ and
$\mathrm{curl}(\mathbf{f}_t)=(\Delta_x v-\Delta_y u)/s$,
and report their squared energy:
\begin{equation}
  \mathrm{DivCurlE}
  =
  \frac{1}{T-1}\sum_{t=1}^{T-1}
  \mathbb{E}_{i,j}\!\left[
    \mathrm{div}(\mathbf{f}_t)_{i,j}^{2}+\mathrm{curl}(\mathbf{f}_t)_{i,j}^{2}
  \right],
\end{equation}
where the expectation averages over valid cells (visibility-masked). Lower DivCurlE indicates less high-frequency spatial noise in the predicted motion.

\noindent (iv) \textit{Visibility-masked Endpoint Error (VEPE).}
We evaluate \vaename{} reconstruction fidelity with endpoint error in pixel coordinates, averaged over visible points.
Let $\mathbf{p}_{i,n}\in\mathbb{R}^2$ be the ground-truth location and $\hat{\mathbf{p}}_{i,n}\in\mathbb{R}^2$ the reconstructed location.
The per-point error is $ e_{i,n} = \left\|\hat{\mathbf{p}}_{i,n}-\mathbf{p}_{i,n}\right\|_2$,
and we average over visible points using $v_{i,n}\in\{0,1\}$:
\begin{equation}
  \mathrm{VEPE}
  =
  \frac{\sum_{i,n} v_{i,n}\, e_{i,n}}{\sum_{i,n} v_{i,n}}.
\end{equation}

\subsection{Quantitative Results.}
\label{sec:quant}
\begin{table}[t]
\centering
\caption{Comparison of trajectory generation with WHN (L).
FVMD and flow diagnostics (FlowTV $\times 10^{2}$, DivCurlE $\times 10^{3}$; $\downarrow$) on Kinetics, RoboTAP, Kubric (MOVi-A; same config as WHN), and MagicData (E).}
\setlength{\tabcolsep}{3pt}
\renewcommand{\arraystretch}{1.05}
\begin{tabular}{l|l|c c c c}
\toprule
\textbf{Metric} & \textbf{Method} & \textbf{Kinetics} & \textbf{RoboTAP} & \textbf{Kubric} & \textbf{MagicData (E)} \\
\midrule

\multirow{2}{*}{FVMD $\downarrow$}
& WHN (L)        & 8999 & 7587 & 4872 & 10383 \\
& \textbf{Ours}  & \textbf{3626} & \textbf{2467} & \textbf{1338} & \textbf{2968} \\
\midrule

\multirow{2}{*}{FlowTV ($10^{2}$) $\downarrow$}
& WHN (L)        & 24.71 & 27.26 & 12.56 & 26.74 \\
& \textbf{Ours}           & \textbf{7.53}  & \textbf{6.03}  & \textbf{6.03}  & \textbf{7.11} \\
\midrule

\multirow{2}{*}{DivCurlE ($10^{3}$) $\downarrow$}
& WHN (L)        & 34.69 & 37.39 & 9.89 & 40.04 \\
& \textbf{Ours}           & \textbf{3.97}  & \textbf{2.03}  & \textbf{2.09} & \textbf{2.99} \\
\bottomrule
\end{tabular}
\label{tab:flow_metrics_whn_vs_ours_nostar}
\end{table}
\noindent \textbf{Future-trajectory generation.}
Table~\ref{tab:flow_metrics_whn_vs_ours_nostar} compares our generator and WHN (L) on real data (Kinetics, RoboTAP), synthetic data (Kubric), and MagicData (E). Kubric uses the same MOVi-A setup as WHN (L); we re-render the dataset for direct comparison. Our method consistently improves motion quality, reducing FVMD by $2.5$--$3.6\times$ (e.g., 4872 to 1338 on Kubric). It also lowers FlowTV and DivCurlE, indicating fewer spatial discontinuities and more stable motion.

\noindent \textbf{Track-VAE reconstruction.}
Table~\ref{tab:vae_recon_main} reports trajectory VAE reconstruction fidelity. \vaename{} achieves low VEPE and outperforms WHN (L)-VAE by a large margin on all datasets. Performance remains stable as segment length increases from 24 to 81 frames, suggesting that the latent representation preserves long temporal windows without accumulating drift.

\begin{table}[th]
\centering
\caption{Fidelity of trajectory VAE reconstruction.
Visibility masked endpoint error (VEPE, ($\downarrow$)) for reconstructing 24 and 81 frame trajectory segments.}
\setlength{\tabcolsep}{4pt}
\renewcommand{\arraystretch}{1.05}
\begin{tabular}{c l|c c c c}
\toprule
\textbf{Frames} & \textbf{Method} &
\textbf{Kinetics} & \textbf{RoboTAP} & \textbf{Kubric} & \textbf{MagicData (E)} \\
\midrule
\multirow{2}{*}{24}
& WHN (L)-VAE          & 63.09 & 29.35 & 15.97 & 69.88 \\
& \textbf{Ours}        & \textbf{1.99} & \textbf{1.59} & \textbf{0.75} & \textbf{1.72} \\
\midrule
\multirow{2}{*}{81}
& WHN (L)-VAE          & 63.31 & 31.52 & 16.27 & 67.97 \\
& \textbf{Ours}        & \textbf{2.04} & \textbf{1.61} & \textbf{0.77} & \textbf{1.72} \\
\bottomrule
\end{tabular}
\label{tab:vae_recon_main}
\end{table}

\subsection{Qualitative Results}
\label{sec:qual}
\begin{figure}[ht]
    \centering
    \includegraphics[width=\linewidth]{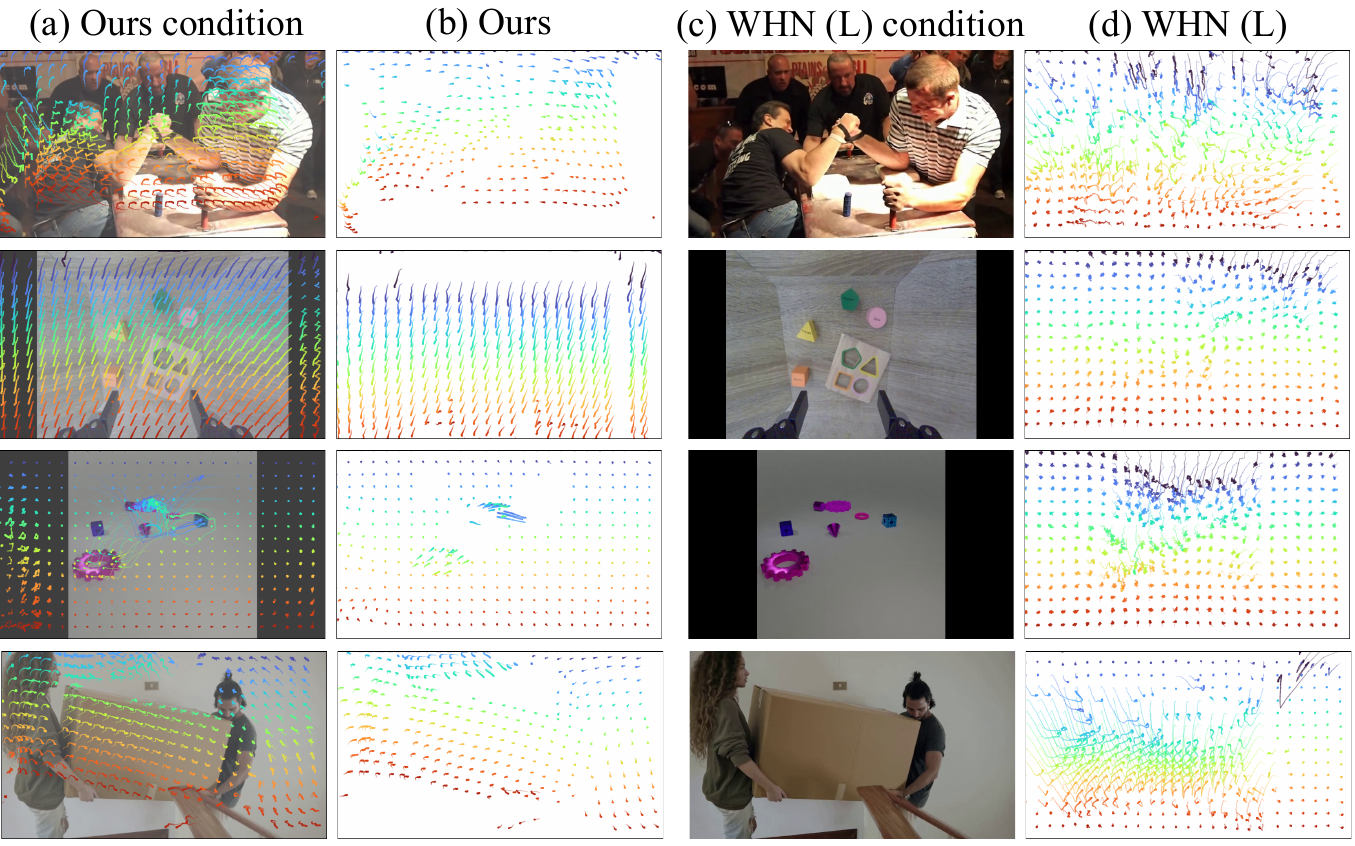}
\caption{Comparison with WHN (L). Each row shows a dataset: Kinetics, RoboTAP, Kubric, and MagicData (E). Our method yields smoother and more coherent motion.}
\label{fig:qual_main}
\end{figure}
\noindent Figure~\ref{fig:qual_main} compares future trajectories across all benchmark sources. Conditioning on observed motion history produces futures that better continue current dynamics and remain spatially coherent, whereas WHN (L) more often shows drift and spatial tearing. More examples are provided in Appendix~\ref{app:more-quali}.

\subsection{Ablation Study}
\label{sec:ablation}
Table~\ref{tab:flow_metrics_ablation_offset} shows the effect of \encodingname{} for 81-frame future generation. Removing offsets degrades long-horizon motion, increasing FVMD-Long and worsening FlowTV and DivCurlE, indicating more drift and tearing. We also ablate \vaename{} offset encoding and the spatiotemporal consistency regularizer, both of which improve reconstruction (Appendix~\ref{app:vae_ablation}). Ablations of \modelname{} boundary hints are in Appendix~\ref{app:generator_ablation}.

\begin{table}[!t]
\centering
\caption{Ablation study with \encodingname{} on future 81 frames.}
\setlength{\tabcolsep}{2pt}
\renewcommand{\arraystretch}{1.05}
\begin{tabular}{l|l|c c c c}
\toprule
\textbf{Metric} & \textbf{Method} & \textbf{Kinetics} & \textbf{RoboTAP} & \textbf{Kubric} & \textbf{MagicData (E)} \\
\midrule

\multirow{2}{*}{FVMD-Long ($10^{-3}$) $\downarrow$}
& w/o offset        & 46.61 & 36.43 & 31.66 & 41.97 \\
& \textbf{Ours}     & \textbf{35.06} & \textbf{21.75} &  \textbf{7.24} & \textbf{28.27} \\
\midrule

\multirow{2}{*}{FlowTV ($10^{2}$) $\downarrow$}
& w/o offset        & 15.92 & 15.26 & 15.54 & 15.01 \\
& \textbf{Ours}     &  \textbf{7.65} &  \textbf{5.96} &  \textbf{5.03} &  \textbf{7.11} \\
\midrule

\multirow{2}{*}{DivCurlE ($10^{3}$) $\downarrow$}
& w/o offset        & 19.10 & 16.27 & 17.66 & 16.13 \\
& \textbf{Ours}     &  \textbf{4.21} &  \textbf{1.89} &  \textbf{1.45} &  \textbf{3.09} \\
\bottomrule
\end{tabular}
\label{tab:flow_metrics_ablation_offset}
\end{table}

\subsection{Downstream Applications}
\label{sec:downstream}
\begin{figure}[t]
    \centering
    \includegraphics[width=\linewidth]{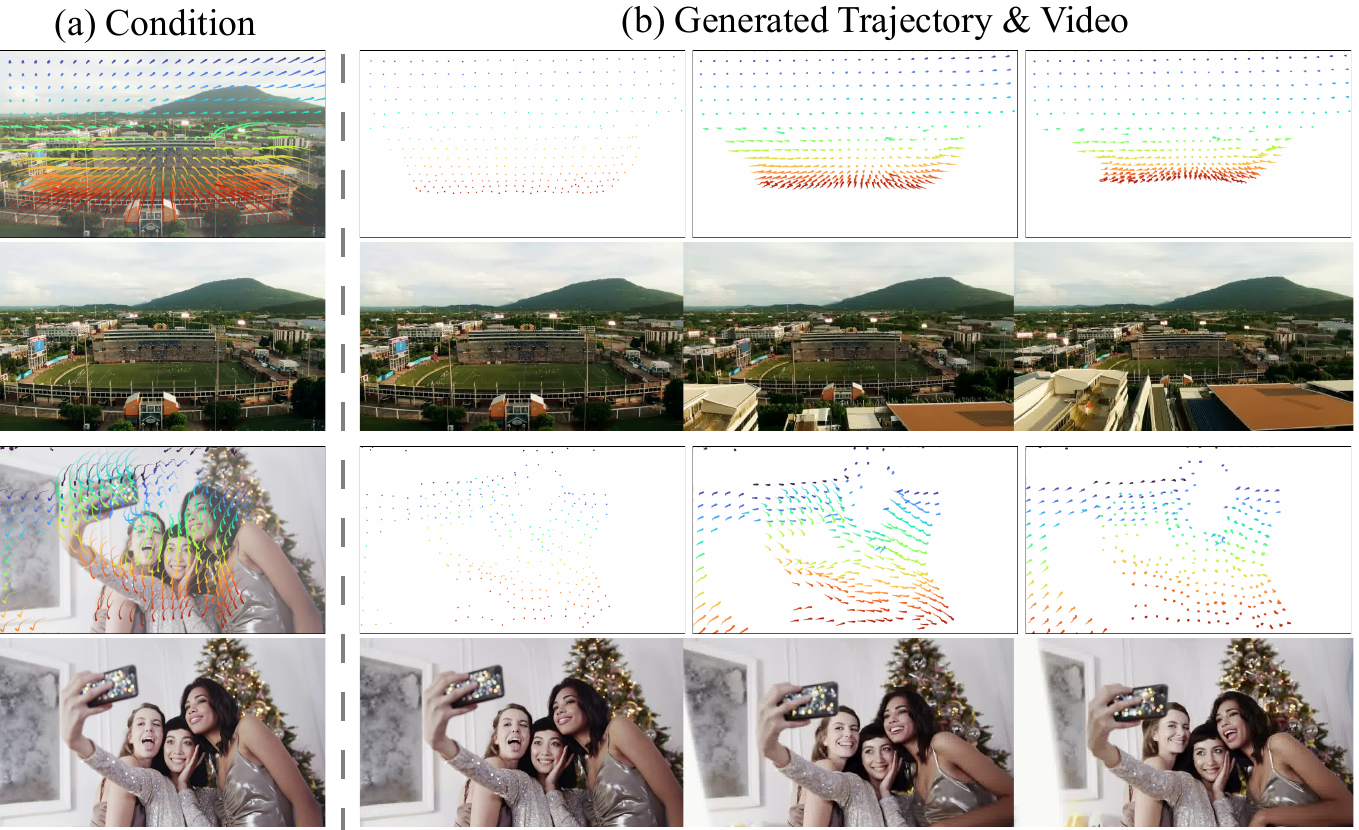}
\caption{Trajectory-guided video generation with Wan-Move~\cite{chu2025wanmove}. We use the observed history to generate a trajectory. Wan-Move then uses this trajectory, along with the condition image, to generate 81 frames, as shown in the third and fourth rows.}
\label{fig:wanmove}
\end{figure}

Predicted future trajectories can serve as motion-control signals for video generation with Wan-Move~\cite{chu2025wanmove}. As shown in Figure~\ref{fig:wanmove}, we use observed history trajectories to generate future trajectory fields with \modelname{}, then feed them to Wan-Move with a single input image to synthesize motion-consistent videos. This demonstrates that trajectory generation enables controllable motion continuation in downstream video generation.


\section{Conclusion}
\noindent In this work, we present a framework for dense future-trajectory generation from observed history. Our approach combines \encodingname{}, a consistency-regularized \vaename{}, and the rectified-flow generator \modelname{} with explicit boundary cues and on-policy fine-tuning. Together, these components produce stable long-horizon motion and outperform state-of-the-art methods on \benchmarkname{} in both quantitative and qualitative evaluations. For future work, we plan to improve trajectory controllability (\eg, user-driven trajectory editing) and further integrate our model with motion-guided generation and editing methods to improve versatility and accuracy.

\section*{Acknowledgements}
This work was supported, in part, by the NSERC DG Grant (No. RGPIN-2022-04636), the NSERC Alliance Grant (ALLRP 604621-25), the Vector Institute for AI, and the Canada CIFAR AI Chair program. 
Resources used in preparing this research were provided, in part, by the Province of Ontario, the Government of Canada through the Digital Research Alliance of Canada \url{alliance.can.ca}, and companies sponsoring the Vector Institute \url{www.vectorinstitute.ai/\#partners}, and Advanced Research Computing at the University of British Columbia. 
Additional resource was provided by the Canada Foundation for Innovation (CFI) via the John R. Evans Leaders Fund (JELF).

%
%

\newpage

\bibliographystyle{plainnat}
\bibliography{main_text/main}

\newpage
\appendix

\section{Model and Training Details}
\label{app:train-model}

\paragraph{Backbone.}
Following WHN~\cite{boduljak2026what}, both \vaename{} and \modelname{} use a DiT-style transformer~\cite{peebles2023scalable} with the Latte design~\cite{ma2025latte}. We operate on trajectory fields at $480{\times}832$ with patch size $32$, yielding $15{\times}26{=}390$ spatial tokens per frame.

\subsection{\vaename{}}

\paragraph{Architecture.}
\vaename{} encodes 81-frame trajectory segments (2D offsets; Section~\ref{sec:encoding}) into a compact spatiotemporal latent tensor with $C{=}16$ channels. We use a Latte encoder/decoder with 16 blocks each, 8 attention heads, and hidden size 512. A temporal convolutional compression module with stride 4 reduces the temporal length from 81 to $21$ latent steps.

\paragraph{Objective.}
We train \vaename{} with a visibility-masked Huber reconstruction loss, a KL penalty with $\beta=5{\times}10^{-5}$, and the spatiotemporal consistency regularizer from Section~\ref{sec:vae}. The regularizer weights are $\lambda_{\mathrm{temporal}}=0.1$ and $\lambda_{\mathrm{spatial}}=0.2$. For the multiscale spatial term, we use $\mathcal{S}=\{1,2,4\}$ and
$\alpha_\delta=1/\delta$, i.e.\ $\{1,0.5,0.25\}$ for hop distances $\{1,2,4\}$.

\paragraph{Optimization.}
We optimize with AdamW~\cite{DBLP:journals/corr/KingmaB14,loshchilov2018decoupled} using a learning rate of $2{\times}10^{-5}$, $(\beta_1,\beta_2)=(0.9,0.999)$, weight decay of 0, batch size 4, and gradient accumulation 2 on 2 RTX PRO 6000 (effective batch size 16), using bf16 mixed precision and clipping gradients to keep the norm at 1.0, training for a total of 30k steps.

\subsection{\modelname{}}

\paragraph{Architecture and conditioning.}
\modelname{} predicts future \vaename{} latents with a Latte~\cite{ma2025latte} generator of 16 blocks, 12 heads, and hidden size 768 (matching WHN (L) scale).
In our setting, $\mathbf{c}$ includes history trajectory latents $\mathbf{z}^{p}$,
tokenized history visibility, history-video features, and a pooled text embedding~\cite{2020t5}.

\paragraph{Latent normalization.}
Before training \modelname{}, we compute per-channel mean and std statistics of \vaename{} latents on the training set and apply channel-wise normalization. The generator is trained in the normalized latent space; decoding uses the inverse transform.

\paragraph{Flow-matching pretraining.}
We train with rectified flow matching in latent space (Section~\ref{sec:generator_fm}) using tube noise $\sigma=0.05$. We sample flow time $t\in(0,1)$ from a mixture: with probability $0.2$, $t\sim\mathrm{Uniform}(0,0.1)$; otherwise $t=\sigma(\mathcal{N}(0,1))$ (clamped to $[10^{-5},1-10^{-5}]$)~\cite{polyak2024movie, boduljak2026what}. We supervise with a token-masked MSE using future visibility pooled to the latent grid, and we assign a small weight 0.01 to invisible tokens so they are not completely ignored.

\paragraph{Source distribution ($z_0$).}
We use the boundary hints: $z_0$ is sampled as Gaussian noise, and the first future latent slice is anchored to the last history latent slice (with additive noise of std 0.1) to encourage continuity at the history--future boundary (Section~\ref{sec:generator_hints}).

\paragraph{Optimization.}
We optimize with AdamW (weight decay 0) at a learning rate of $6{\times}10^{-5}$, batch size 32 on 4 H100 with an effective batch size of 128, bf16 mixed precision, and gradient clipping at 1.0. We enable checkpointing for activation to reduce memory usage. And we trained 100k steps in total.

\subsection{On-policy $K$-step fine-tuning}
\label{app:onpolicy}
To reduce the train--test mismatch from ODE sampling (Section~\ref{sec:generator_kstep}), we fine-tune \modelname{} with the on-policy $K$-step rollout objective. We use $K{=}8$ Euler rollout steps on a logit-spaced time grid in $[t_\epsilon,1-t_\epsilon]$ with $t_\epsilon=10^{-5}$, and clamp denominators with $t_\epsilon=10^{-3}$. The rollout loss uses weights $w_1=1.0$ (toward $\mathbf{z}_1$), $w_0=0.5$ (toward $\mathbf{z}_0$), and endpoint-consistency weight $\gamma$ as $0.1$. For stability, the endpoint-consistency term is computed \emph{without} visibility masking (i.e., over all tokens). We apply the rollout loss on a small sub-batch of 8 samples per iteration, with overall weight $ \lambda_{\text{k-step}}=0.1$. And fine-tune using learning rate $1{\times}10^{-5}$ while keeping other settings unchanged.

\section{Further Analyses of Offset Encoding and Consistency Regularization}
This section provides further analyses of two key design choices in our method. We first quantify how \encodingname{} reduce location-dependent bias, and then illustrate why pointwise reconstruction alone is insufficient without spatiotemporal consistency regularization.
\subsection{Quantifying Location Bias Removed by Grid-Anchored Offsets}
\label{app:offset_insight}

This appendix provides details for the model-free analysis in Figure~\ref{fig:offset-insight-metric}.
The goal is to measure how much of the coordinate variance is explained purely by the \emph{grid location} (a static spatial baseline), compared to the remaining variance that comes from \emph{temporal motion}.

We use dense point tracks extracted by AllTracker~\cite{harley2025alltracker} on a stride-$s$ grid (here $s=32$) at resolution $H\times W=480\times 832$. 
Each track corresponds to a fixed grid cell $n\in\{1,\dots,N\}$, and provides per-frame 2D coordinates and visibility.
We convert pixel coordinates to normalized coordinates in $[-1,1]$ and form offsets by subtracting the (normalized) center anchor of the corresponding stride cell, i.e.\ $\mathbf{X}_{n,t}=\mathbf{D}_{n,t}-\mathbf{G}_n$.
We evaluate on 128 randomly sampled clips.

For a single coordinate axis $d$ (either $D_x$, $D_y$, $X_x$, or $X_y$), let $d_{n,t}$ be the value at grid location $n$ and time $t$, with visibility $v_{n,t}\in\{0,1\}$.
For each location $n$, we compute a visibility-weighted time average
$\mu_n = \frac{\sum_t v_{n,t}\, d_{n,t}}{\sum_t v_{n,t}}$
as the \emph{location baseline}, and a visibility-weighted temporal variance around that baseline
$\sigma_n^2 = \frac{\sum_t v_{n,t}\,(d_{n,t}-\mu_n)^2}{\sum_t v_{n,t}}$.

We then decompose the total coordinate variance into a between-location term and a within-location term.
Let $\mathrm{Var}_n(\cdot)$ and $\mathbb{E}_n[\cdot]$ denote variance and mean over grid locations $n$.
We define the fraction of variance explained by grid location as:
\begin{equation}
\mathrm{Var}(d)
=
\mathrm{Var}_{n}(\mu_n)
+
\mathbb{E}_{n}\!\left[\sigma_n^2\right],
\quad
\mathrm{Explained}(\%)
=
100\cdot
\frac{\mathrm{Var}_{n}(\mu_n)}
{\mathrm{Var}_{n}(\mu_n)+\mathbb{E}_{n}\!\left[\sigma_n^2\right]}.
\label{eq:offset_explained_var}
\end{equation}
Here, $\mathrm{Var}(d)$ denotes the total variance under this decomposition: $\mathrm{Var}_{n}(\mu_n)$ measures how much coordinates differ across grid locations due to the static spatial baseline, while $\mathbb{E}_{n}[\sigma_n^2]$ measures the average temporal variance (motion) at a fixed location.
A high explained percentage indicates that most coordinate variation is attributable to grid position rather than motion.
As shown in Figure~\ref{fig:offset-insight-metric}, absolute coordinates have high explained variance, while offsets strongly reduce this location-driven component.

\subsection{Why spatiotemporal consistency regularization is needed}
\label{app:vae_toy}
In this section, we explain why pointwise reconstruction loss alone is insufficient and motivate the need for spatiotemporal consistency regularization.
Pointwise reconstruction alone does not uniquely determine a physically plausible motion trajectory.
To illustrate this, we consider a 1D trajectory with full visibility and ground truth $x(t)=t$ for discrete time $t$.
We compare two reconstructions that have the same per-frame deviation magnitude: a \emph{smooth} reconstruction with a constant bias,
$\hat{x}_{\text{smooth}}(t)=t+b$, and a \emph{jittery} reconstruction with an alternating bias,
$\hat{x}_{\text{jitter}}(t)=t+b(-1)^t$.
Both reconstructions have identical pointwise error $|\hat{x}(t)-x(t)|=|b|$ at every frame, and therefore can achieve the same masked Huber reconstruction loss $L_{\mathrm{rec}}$ (for any Huber threshold that treats these deviations similarly).

\begin{figure}[thb]
  \centering
  \begin{subfigure}[t]{0.49\linewidth}
    \centering
    \includegraphics[width=\linewidth]{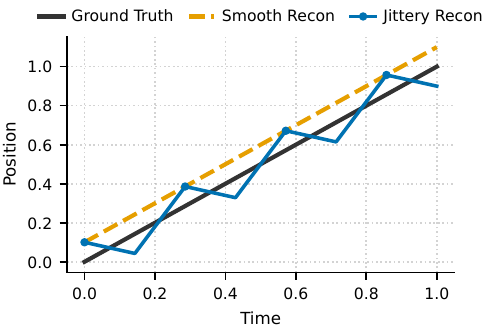}
    \caption{Ground truth, smooth, and jittery.}
    \label{fig:short-a}
  \end{subfigure}\hfill
  \begin{subfigure}[t]{0.49\linewidth}
    \centering
    \includegraphics[width=\linewidth]{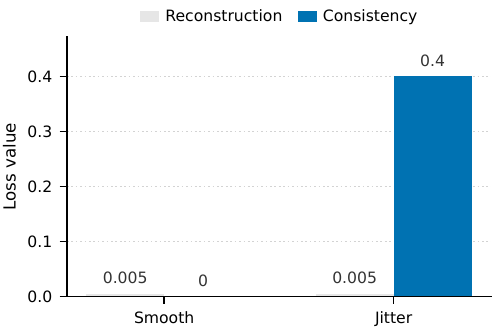}
    \caption{Same $L_{\mathrm{rec}}$, different $L_{\mathrm{st}}$.}
    \label{fig:short-b}
  \end{subfigure}
  \caption{1D toy example: pointwise reconstruction loss does not guarantee realistic motion. Both smooth and jittery reconstructions of $x(t)=t$ can match $L_{\mathrm{rec}}$, but only the smooth one aligns with ground truth motion. Consistency regularization separates realistic from unrealistic solutions.}
  \label{fig:vae-toy-jitter}
\end{figure}

However, the two reconstructions imply very different \emph{motion}.
The ground-truth velocity is constant, $\Delta x(t)=x(t)-x(t-1)=1$.
The smooth reconstruction preserves this, $\Delta \hat{x}_{\text{smooth}}(t)=1$, while the jittery reconstruction alternates its frame-to-frame displacement,
$\Delta \hat{x}_{\text{jitter}}(t)=1+2b(-1)^t$.
As a result, the velocity-matching term penalizes the jittery solution even though $L_{\mathrm{rec}}$ cannot distinguish it. The same issue generalizes to dense 2D trajectory fields: reconstructions can match per-frame positions yet exhibit temporal jitter and locally inconsistent motion across neighboring points. Our spatiotemporal consistency regularizer directly targets these failure modes by matching temporal displacements ($L_{\mathrm{temporal}}$) and neighbor relations ($L_{\mathrm{spatial}}$), as shown in Figure~\ref{fig:vae-toy-jitter}.

\section{Additional Qualitative Results}
\label{app:app_qua}

\subsection{More Comparisons with WHN}
\label{app:more-quali}
\begin{figure}[!t]
    \centering
    \includegraphics[width=\linewidth]{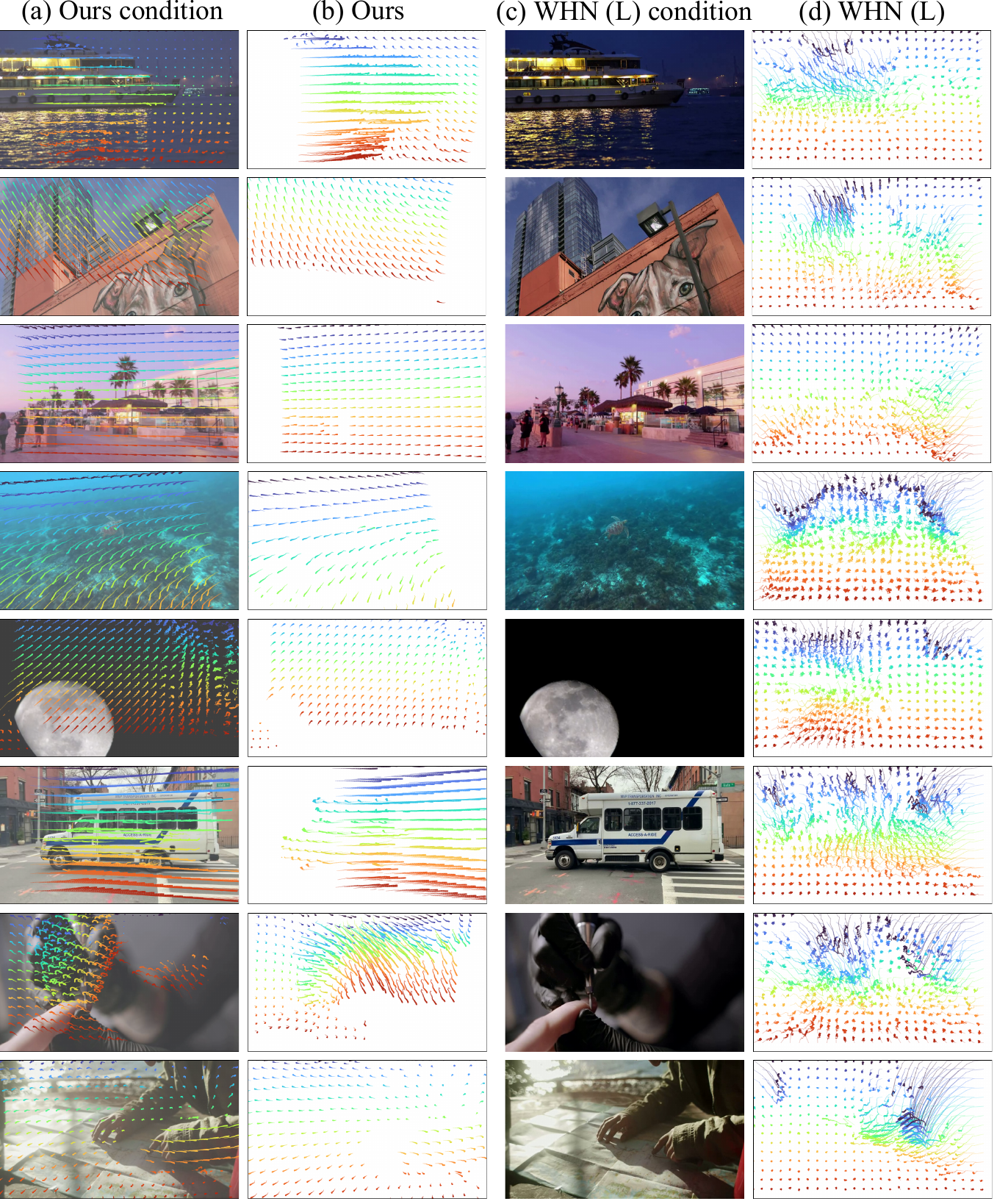}
\caption{More Comparison with WHN (L). }
\label{fig:qual_2}
\end{figure}
\noindent Figure~\ref{fig:qual_2} shows additional side-by-side comparisons with WHN (L).
Each row is a different clip; (a)(b) show our history condition and predicted future, while (c)(d) show WHN (L) with its image condition and predicted future. Our results more consistently preserve long-horizon coherence with fewer drift/tearing artifacts.

\subsection{More Downstream Examples}
\label{app:downstream}
\begin{figure}[!t]
    \centering
    \includegraphics[width=\linewidth]{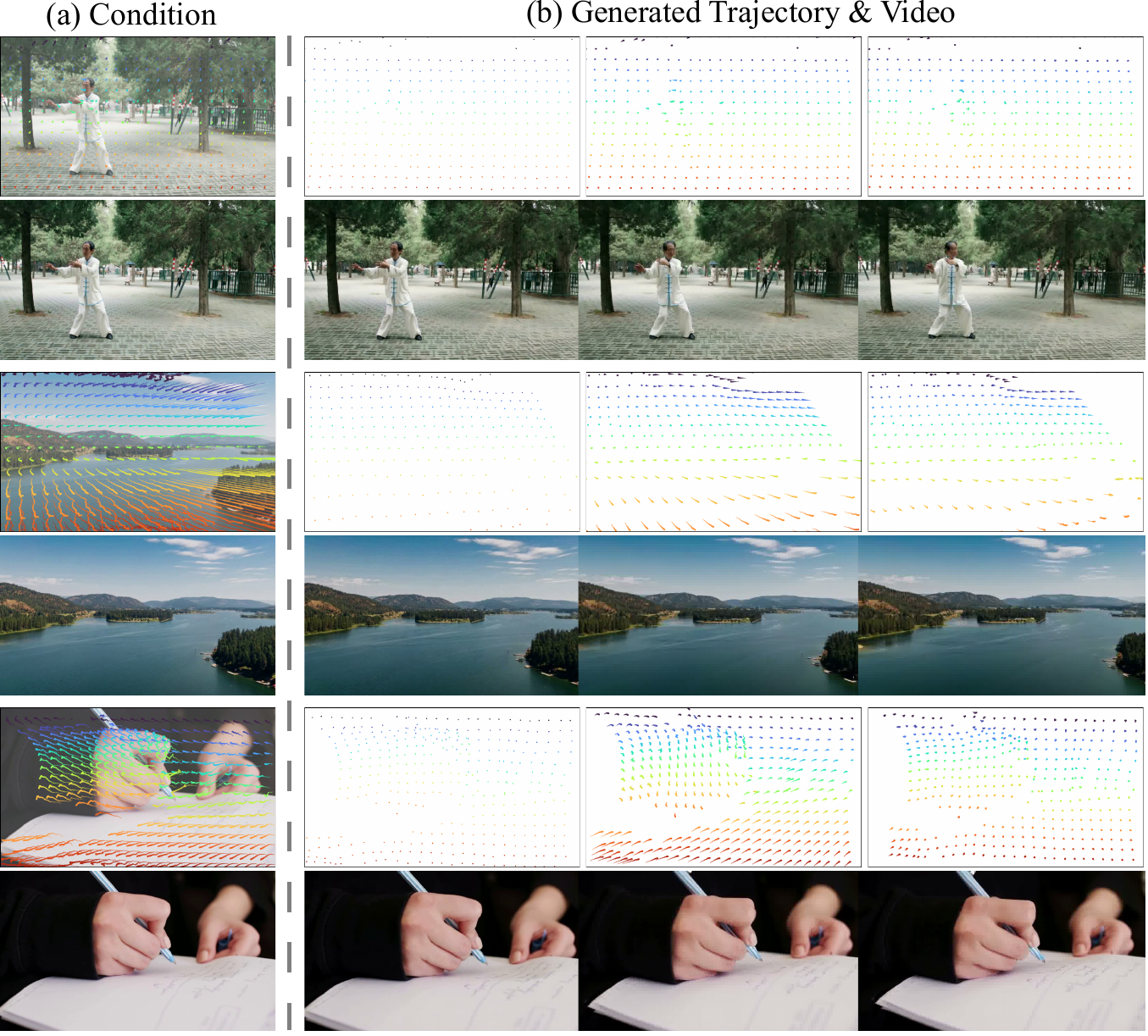}
\caption{Additional Wan-Move results driven by our predicted trajectories. Left: conditioning inputs (history trajectory and the conditioning image). Right: our generated 81-frame future trajectories and the corresponding 81-frame videos synthesized by Wan-Move.}
\label{fig:wanmove2}
\end{figure}
We further demonstrate downstream controllable video synthesis using Wan-Move~\cite{chu2025wanmove}.
Given the observed history, we first sample 81-frame future trajectories with \modelname{}.
We then provide Wan-Move with the predicted trajectories and the conditioning image to generate an 81-frame video that follows the generated motion.
Figure~\ref{fig:wanmove2} shows additional examples.

\section{Additional Quantitative Studies}
This section presents additional empirical results beyond the main paper. We first report ablations and training dynamics for \vaename{}, and then provide component ablations and sampling analyses for \modelname{}.

\subsection{\vaename{} Ablations}
\label{app:vae_ablation}

We ablate two components of \vaename{}: (i) the \encodingname{} and (ii) the spatiotemporal consistency regularizer. Table~\ref{tab:vae_ablation} reports VEPE (pixels; lower is better) for reconstructing 24- and 81-frame trajectory segments. Removing offset encoding consistently increases reconstruction error across datasets and horizons, confirming that offsets reduce location-dependent bias and simplify trajectory modeling. Removing the consistency regularizer also degrades VEPE, especially on longer windows, consistent with the regularizer suppressing temporal jitter and improving local coherence in the reconstructed motion.

\begin{table}[!t]
\centering
\caption{\vaename{} ablations.
Effect of \encodingname{} and spatiotemporal consistency regularization on VEPE ($\downarrow$).}
\setlength{\tabcolsep}{4pt}
\renewcommand{\arraystretch}{1.05}
\begin{tabular}{c l|c c c c}
\toprule
\textbf{Frames} & \textbf{Method} &
\textbf{Kinetics} & \textbf{RoboTAP} & \textbf{Kubric} & \textbf{MagicData (E)} \\
\midrule
\multirow{3}{*}{24}
& w/o offset           & 4.24 & 3.98 & 3.26 & 3.91 \\
& w/o regularizer      & 2.99 & 2.35 & 1.42 & 2.65 \\
& \textbf{Ours}        & \textbf{1.99} & \textbf{1.59} & \textbf{0.75} & \textbf{1.72} \\
\midrule
\multirow{3}{*}{81}
& w/o offset           & 4.30 & 4.01 & 3.24 & 3.99 \\
& w/o regularizer      & 3.11 & 2.41 & 1.47 & 3.99 \\
& \textbf{Ours}        & \textbf{2.04} & \textbf{1.61} & \textbf{0.77} & \textbf{1.72} \\
\bottomrule
\end{tabular}
\label{tab:vae_ablation}
\end{table}

\subsection{VAE Training Dynamics}
\label{app:vae-train}
\begin{figure}[htb]
    \centering
    \includegraphics[width=0.6\linewidth]{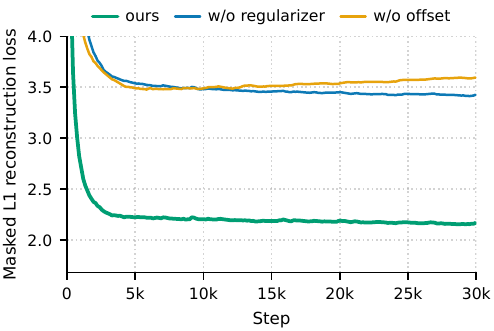}
\caption{\vaename{} training curves. Masked $L_1$ reconstruction metric vs. training step. Using \encodingname{} and the spatiotemporal regularizer yields faster, more stable convergence and a substantially lower loss than removing offsets or the regularizer.}
\label{fig:vae_train_loss}
\end{figure}
\noindent Figure~\ref{fig:vae_train_loss} plots the \vaename{} masked $L_1$ reconstruction loss for our full model and two ablations. Removing the \encodingname{} leads to worse convergence and a gradually increasing loss, consistent with overfitting to location-dependent coordinate biases in absolute space. In contrast, removing the spatiotemporal consistency regularizer yields a higher-loss plateau, indicating underfitting: pointwise reconstruction alone does not sufficiently constrain motion structure. Together, these trends support our design: offset encoding makes the learning problem better conditioned by removing global-position bias, while the regularizer provides motion-level constraints that help the latent space capture temporally smooth and locally coherent trajectories.

\subsection{\modelname{} Ablation Study}
\label{app:generator_ablation}
This appendix provides additional ablations on \modelname{} components described in Section~\ref{sec:generator_hints}, including boundary hints and on-policy fine-tuning.
All results are computed on 81-frame future prediction using Euler integration with 10 steps, and report the same motion and flow-field diagnostics as in the main paper.

We ablate:
\begin{enumerate}
    \item \textbf{History fusion:} removing the token-aligned history fusion module, so history is provided only through the conditioning stream $\mathbf{c}$.
    \item \textbf{Boundary anchoring:} disabling first-slice boundary anchoring, i.e.\ not adding
$\mathbf{z}(-1,n)$ to $\mathbf{z}_0(0,n)$.
    \item \textbf{On-policy fine-tuning:} training with flow matching only (Section~\ref{sec:generator_fm}), \ie setting $\lambda_{\mathrm{k-step}}=0$ and omitting the on-policy $K$-step fine-tuning stage.
\end{enumerate}

In addition, Table~\ref{tab:flow_metrics_ablation_components} reports an early flow-matching checkpoint (20k steps) and a full-budget model trained without on-policy fine-tuning (w/o on-policy) to separate the effect of training duration from the proposed on-policy stage.
Overall, longer flow-matching training improves motion quality, but on-policy fine-tuning consistently provides further gains across datasets.
Among the boundary-hint variants, disabling first-slice boundary anchoring (w/o anchoring) causes the largest degradation, while removing token-aligned history fusion (w/o fusion) yields a smaller but consistent drop.

\begin{table}[!t]
\centering
\caption{Ablation study with flow-field diagnostics (generator components).
Metrics are computed on 81-frame future prediction. Lower is better ($\downarrow$).
All results use Euler integration with 10 steps.
\textbf{w/o fusion} removes the token-aligned history fusion module;
\textbf{w/o anchoring} disables boundary anchoring of the first future latent slice ;
\textbf{20k steps} is an early flow-matching checkpoint of the generator;
\textbf{w/o on-policy} matches the full flow-matching training budget of our method but omits the on-policy $K$-step fine-tuning stage.}
\setlength{\tabcolsep}{1pt}
\renewcommand{\arraystretch}{1.05}
\begin{tabular}{l|l|c c c c}
\toprule
\textbf{Metric} & \textbf{Setting} & \textbf{Kinetics} & \textbf{RoboTAP} & \textbf{Kubric} & \textbf{MagicData (E)} \\
\midrule

\multirow{5}{*}{FVMD-Long ($10^{-3}$) $\downarrow$}
& w/o fusion            & 39.60 & 28.86 & 16.38 & 33.87 \\
&  w/o anchoring        & 43.50 & 33.12 & 20.29 & 37.11 \\
& 20 k steps                     & 40.13 & 28.81 & 17.02 & 33.51 \\
& w/o on-policy                    & 38.87 & 27.38 & 13.86 & 31.71 \\
& \textbf{Ours}   & \textbf{35.06} & \textbf{21.75} &  \textbf{7.24} & \textbf{28.27} \\
\midrule

\multirow{5}{*}{FlowTV ($10^{2}$) $\downarrow$}
& w/o fusion             &  9.66 &  8.66 &  9.56 &  8.50 \\
& w/o anchoring        & 10.07 &  9.11 &  9.80 &  8.90 \\
& 20 k steps                     &  9.99 &  9.09 &  9.99 &  8.94 \\
& w/o on-policy                    &  8.98 &  8.22 &  8.90 &  8.16 \\
& \textbf{Ours}   &  \textbf{7.65} &  \textbf{5.96} &  \textbf{5.03} &  \textbf{7.11} \\
\midrule

\multirow{5}{*}{DivCurlE ($10^{3}$) $\downarrow$}
& w/o fusion             &  7.29 &  5.49 & 11.24 &  5.04 \\
& w/o anchoring        &  7.83 &  5.85 & 11.71 &  5.53 \\
& 20 k steps                     &  8.09 &  5.79 & 11.44 &  5.63 \\
& w/o on-policy                    &  6.11 &  4.58 &  8.12 &  4.41 \\
& \textbf{Ours}   &  \textbf{4.21} &  \textbf{1.89} &  \textbf{1.45} &  \textbf{3.09} \\
\bottomrule
\end{tabular}
\label{tab:flow_metrics_ablation_components}
\end{table}




\subsection{Advanced ODE Solver and More Sampling Steps}
\label{app:advanced_sampler}

Our generator is sampled by integrating the rectified-flow ODE in the latent space.
In the main paper, we use a lightweight Euler solver with 10 steps as the default for efficiency, to match the WHN setting~\cite{boduljak2026what}.
Here we study the effect of using a higher-order sampler with more function evaluations: we additionally evaluate both WHN (L)~\cite{boduljak2026what} and our method using the Dormand--Prince solver (DOPRI5)~\cite{dormand1980family} with 100 steps, following common practice for improving ODE sampling quality~\cite{tong2024improving}.
We denote results obtained with DOPRI5-100 by a superscript $^{*}$, indicating the use of this solver. Entries without $^{*}$ use the default Euler-10 setting.





\begin{table}[!t]
\centering
\caption{Effect of sampler choice and step count.
Metrics are computed on 81-frame future prediction; lower is better ($\downarrow$).
Entries marked with $^{*}$ use Dormand--Prince (DOPRI5) with 100 steps, while unmarked entries use Euler with 10 steps (default).}
\setlength{\tabcolsep}{3pt}
\renewcommand{\arraystretch}{1.05}
\begin{tabular}{l|l|c c c c}
\toprule
\textbf{Metric} & \textbf{Method} & \textbf{Kinetics} & \textbf{RoboTAP} & \textbf{Kubric} & \textbf{MagicData (E)} \\
\midrule

\multirow{4}{*}{FVMD $\downarrow$}
& WHN (L)                 & 8999 & 7587 & 4872 & 10383 \\
& WHN (L)$^{*}$           & 8755 & 5855 & 6284 &  9691 \\
& \textbf{Ours}           & 3626 & \textbf{2467} & \textbf{1338} & 2968 \\
& \textbf{Ours}$^{*}$     & \textbf{3182} & 2546 & 1724 & \textbf{2358} \\
\midrule

\multirow{4}{*}{FlowTV ($10^{2}$) $\downarrow$}
& WHN (L)                 & 24.71 & 27.26 & 12.56 & 26.74 \\
& WHN (L)$^{*}$           & 22.06 & 14.46 & 18.41 & 14.46 \\
& \textbf{Ours}           &  7.53 &  6.03 &  6.03 &  7.11 \\
& \textbf{Ours}$^{*}$     & \textbf{4.50} & \textbf{3.40} & \textbf{3.09} & \textbf{3.72} \\
\midrule

\multirow{4}{*}{DivCurlE ($10^{3}$) $\downarrow$}
& WHN (L)                 & 34.69 & 37.39 &  9.89 & 40.04 \\
& WHN (L)$^{*}$           & 28.52 & 14.41 & 17.58 & 14.41 \\
& \textbf{Ours}           &  3.97 &  2.03 &  2.09 &  2.99 \\
& \textbf{Ours}$^{*}$     & \textbf{2.86} & \textbf{1.17} & \textbf{1.10} & \textbf{1.59} \\
\bottomrule
\end{tabular}
\label{tab:flow_metrics_whn_vs_ours}
\end{table}

Table~\ref{tab:flow_metrics_whn_vs_ours} shows that increasing solver accuracy can improve sampling quality, especially in the flow-field diagnostics: for our method, DOPRI5-100 consistently reduces FlowTV and DivCurlE across datasets, indicating smoother and more locally coherent predicted motion.

Importantly, our method substantially outperforms WHN (L) under both solver settings, indicating that the gains are not dependent on a particular sampler; we use Euler-10 as a strong efficiency-quality trade-off, and DOPRI5-100 as an optional higher-cost setting when additional smoothness is desired.

\section{Conditioning and Auxiliary Components}
This section describes additional conditioning and auxiliary components used in the full pipeline. We first detail the boundary hints used by \modelname{}, then summarize the camera-motion caption augmentation used to enrich text conditioning, and finally describe the visibility predictor.

\subsection{Boundary Hints Details}
\label{app:boundary}

This appendix specifies the two \emph{boundary hints} used by \modelname{} (Section~\ref{sec:generator_hints}):
(i) a boundary-anchored source state $\mathbf{z}_0$ for rectified-flow sampling, and
(ii) a token-aligned fusion term that injects boundary information from $\mathbf{z}^{p}$ into the model input.

\paragraph{Boundary-anchored source state $\mathbf{z}_0$.}
Let $\mathbf{z}(-1,n)$ denote the last history latent slice.
We initialize the source state by repeating this boundary latent across future latent time indices $k$ and adding Gaussian noise:
\begin{equation}
  \mathbf{z}_0(k,n)=\mathbf{z}(-1,n)+\sigma_0\,\boldsymbol{\eta}(k,n),
  \qquad
  \boldsymbol{\eta}\sim\mathcal{N}(\mathbf{0},\mathbf{I}).
  \label{eq:x0_repeat_last_app}
\end{equation}
In practice, we set $k=0$.

\paragraph{Token-aligned history fusion.}
In addition to conditioning through $\mathbf{c}$, we add a lightweight
\emph{additive} boundary cue to the model's input tokens. Concretely, at any flow time $t$ (including $t=0$ with
$\mathbf{z}_t=\mathbf{z}_0$), we form the token embeddings from $\mathbf{z}_t$ and then add an aligned history term:
\begin{equation}
  \mathrm{Tok}(\mathbf{z}_t)(k,n)
  \leftarrow
  \mathrm{Tok}(\mathbf{z}_t)(k,n)
  +
  \alpha\,g_k\Bigl(\mathbf{b}(n) + \omega_k\,\mathbf{d}(n)\Bigr),
  \label{eq:track_fusion_app}
\end{equation}
where $\mathrm{Tok}(\cdot)$ denotes the model input tokenization, $\alpha$ is a learnable scalar, and
$g_k\in(0,1)$ is an optional learnable per-step gate. Let $\mathbf{z}(-2,n)$ denote the second last history latent slice. The boundary feature $\mathbf{b}(n)$ and velocity hint
$\mathbf{d}(n)$ are computed from the last history steps at the \emph{same} spatial token $n$:
\begin{equation}
  \mathbf{b}(n)=\mathrm{Tok}(\mathbf{z}(-1, n),\qquad
  \mathbf{d}(n)=\mathrm{Tok}(\mathbf{z}(-1, n)-\mathrm{Tok}(\mathbf{z}(-2, n),
\end{equation}
and $\omega_k$ linearly increases from $0$ (first future step) to $1$ (last future step), so the fusion progressively extends the cue deeper into the future horizon.

\subsection{Camera Movement Caption Dataset Augmentation}
\label{app:camera}

To enrich text conditioning with explicit camera cues, we augment MagicData captions with short camera-motion phrases estimated from the extracted point tracks. For each clip, we robustly estimate global camera translation from track displacements and further decompose residual motion into radial (zoom) and tangential (roll) components around the image center; remaining residual jitter is summarized as a handheld/shake score. We then map these statistics to a small vocabulary of camera primitives (e.g., pan/tilt, zoom in/out, static/handheld) with coarse speed buckets, and append the resulting phrase to the original caption during training.

\subsection{Visibility Predictor}
\label{app:vis_pred}
As an auxiliary component, we predict token-level visibility for the generated future motion, since points may become occluded or leave the image.
Given future latent tokens $\mathbf{z}^{f}\in\mathbb{R}^{T\times N\times C}$, we train a lightweight predictor $f_\omega$ that outputs per-token visibility logits
$\boldsymbol{\ell}\in\mathbb{R}^{T\times N}$.

The predictor is applied independently to each spatial token.
For each token, we first apply a linear projection to the $C$-dimensional features, then use a short stack of 1D temporal convolutions along the latent-time axis, and finally apply a $1\times1$ projection to produce one logit per latent time step.
For supervision, we downsample the dense future visibility mask $\mathbf{M}^{f}$ to the \vaename{} token grid using max-pooling (equivalently, a logical OR) to obtain token-level targets.
We optimize a binary cross-entropy loss with logits.

At inference time, we run $f_\omega$ on generated future latents $\hat{\mathbf{z}}^{f}$ and threshold $\sigma(\boldsymbol{\ell})$ to obtain predicted visibility $\hat{\mathbf{m}}^{f}$.
This visibility head is lightweight and independent of the trajectory generator.

\end{document}